\documentclass[journal]{IEEEtran}

\usepackage{cite}
\usepackage{amsmath}
\usepackage{multirow}
\usepackage{graphicx}
\usepackage{subfigure}
\usepackage{caption}
\usepackage{color}
\usepackage{soul}
\usepackage{url}
\usepackage{url}

\soulregister\cite7
\soulregister\ref7

\ifCLASSINFOpdf

\else

\fi

\begin{document}

\title{DRFN: Deep Recurrent Fusion Network for \\Single-Image Super-Resolution with Large Factors}

\author{Xin Yang, Haiyang Mei, Jiqing Zhang, Ke Xu, Baocai Yin, Qiang Zhang, and Xiaopeng Wei, \IEEEmembership{Member, IEEE}
\thanks{This work was supported in part by the National Natural Science Foundation of China (NSFC) under Grant 91748104, Grant 61632006, Grant 61425002, and Grant U1708263. \emph{(Corresponding authors: Baocai Yin, Qiang Zhang, and Xiaopeng Wei.)}}
\thanks{X. Yang, H. Mei, J. Zhang, K. Xu, B. Yin, Q. Zhang, and X. Wei are with the Department of Electronic Information and Electrical Engineering, Dalian University of Technology, Dalian 116024, China (e-mail: \protect\url{xinyang@dlut.edu.cn}; \protect\url{mhy845879017@gmail.com}; \protect\url{zhangjiqing@mail.dlut.edu.cn}; \protect\url{kkangwing@mail.dlut.edu.cn}; \protect\url{ybc@dlut.edu.cn}; \protect\url{zhangq@dlut.edu.cn}; \protect\url{xpwei@dlut.edu.cn}).}
\thanks{Color versions of one or more of the figures in this paper are available online at http://ieeexplore.ieee.org.}
\thanks{The source code can be found at: \protect\url{https://github.com/Mhaiyang/DRFN/}.}
}

\markboth{IEEE TRANSACTIONS ON MULTIMEDIA}%
{Shell \MakeLowercase{\textit{et al.}}: DRFN: Deep Recurrent Fusion Network
for \\Single-Image Super-Resolution with Large Factors}


\maketitle

\begin{abstract}
Recently, single-image super-resolution has made great progress owing to the development of deep convolutional neural networks (CNNs). The vast majority of CNN-based models use a pre-defined upsampling operator, such as bicubic interpolation, to upscale input low-resolution images to the desired size and learn non-linear mapping between the interpolated image and ground truth high-resolution (HR) image. However, interpolation processing can lead to visual artifacts as details are over-smoothed, particularly when the super-resolution factor is high. In this paper, we propose a Deep Recurrent Fusion Network (DRFN), which utilizes transposed convolution instead of bicubic interpolation for upsampling and integrates different-level features extracted from
recurrent residual blocks to reconstruct the final HR images. We adopt a deep recurrence learning strategy and thus have a larger receptive field, which is conducive to reconstructing an image more accurately. Furthermore, we show that the multi-level fusion structure is suitable for dealing with image super-resolution problems. Extensive benchmark evaluations demonstrate that the proposed DRFN performs better than most current deep learning methods in terms of accuracy and visual effects, especially for large-scale images, while using fewer parameters.
\end{abstract}


% \fi
\IEEEpeerreviewmaketitle

\section{Introduction}
\IEEEPARstart{S}{ingle}-image super-resolution (SISR) refers to the transformation of an image from low-resolution (LR) to high-resolution (HR). SISR is a long-standing problem in computer graphics and vision. Higher-resolution images often provide more desired information and can be applied in many domains, such as security and surveillance imaging, medical imaging, satellite imaging, and other fields. Therefore, it is necessary to explore the reconstruction performance of image super-resolution with larger upscaling factors.

Various algorithms have been introduced to solve the super-resolution (SR) problem, beginning with initial work by Freeman et al. \cite{freeman2002example}. Currently, deep-learning-based methods, especially convolutional neural networks (CNNs), are widely used to handle image SR owing to the powerful learning ability of CNNs. Super-Resolution Convolutional Neural Network (SRCNN) \cite{dong2016image} pioneered the use of three-layer CNNs to learn the mapping relationship between an interpolated image and HR image and significantly outperformed traditional non-deep learning methods. After that, Kumar et al. \cite{kumar2016fast} tapped into the ability of polynomial neural networks to hierarchically learn refinements of a function that maps LR to HR patches. Shi et al. \cite{shi2017structure} developed a contextualized multitask learning framework to address the SR problem. Kim et al. proposed two neural network structures with 20-layer convolutions, termed VDSR \cite{kim2016accurate} and DRCN \cite{kim2016deeply} respectively, and achieved state-of-the-art performance. Lim et al. built a wide-network EDSR \cite{Lim_2017_CVPR_Workshops} using residual blocks. To generate photo-realistic natural images, Ledig et al. \cite{ledig2016photo-realistic} presented a generative adversarial network for SR. Lai et al. \cite{LapSRN} proposed a deep convolutional network within a Laplacian pyramid framework, which progressively predicts high-frequency residuals in a coarse-to-fine manner.

\begin{figure}[tbp]
  \centering
  \includegraphics[width = 1\linewidth]{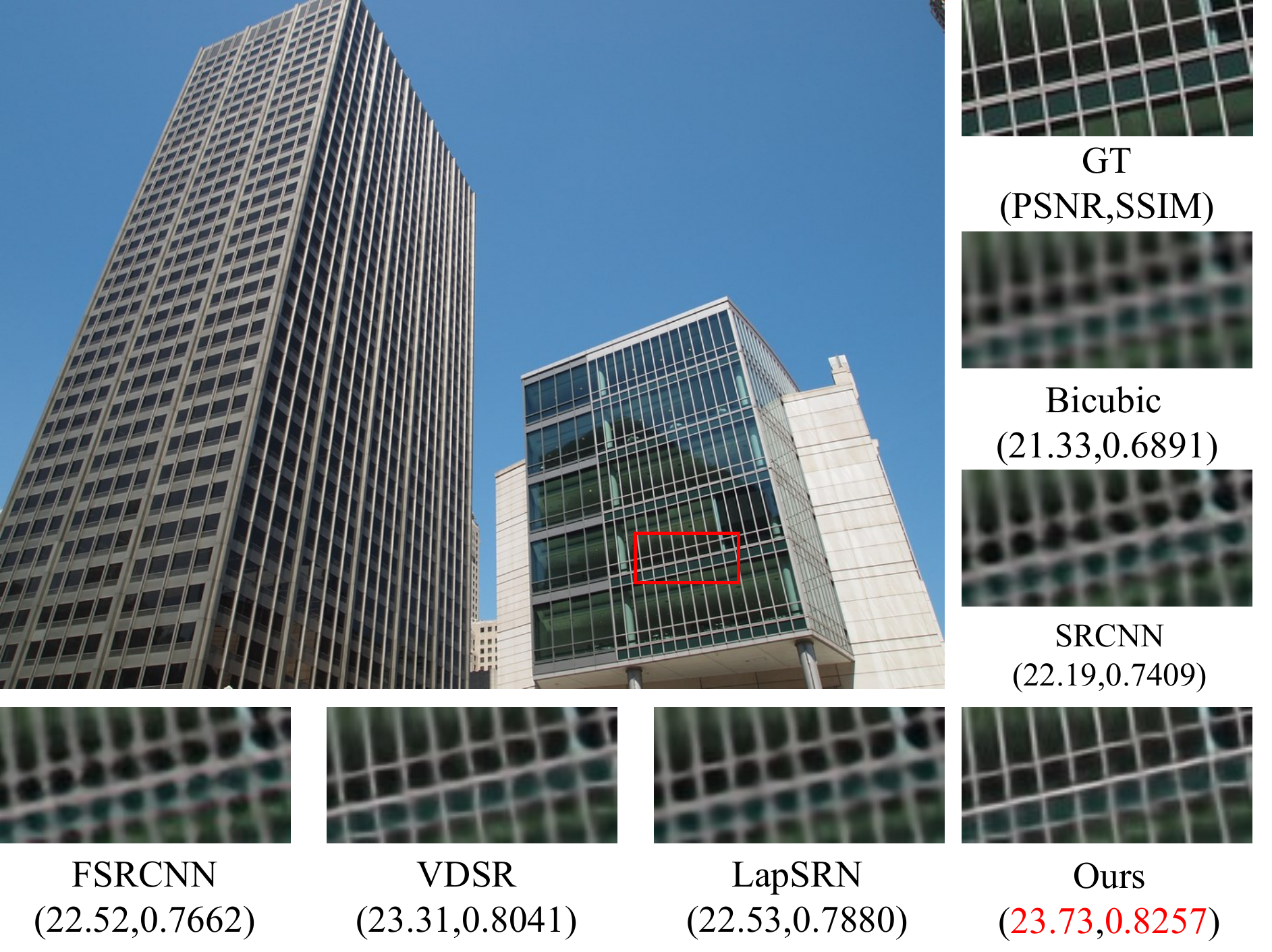}
  \caption{\label{fig:xiaoguo}
            \footnotesize{Visual comparisons of $\times$4 super-resolution on a challenging image from Urban100 \cite{huang2015single}. Results of other methods have serious artifacts and are highly blurred. The proposed method suppresses artifacts effectively and generates clear texture details.}}
\end{figure}

\begin{figure*}[tbp]
  \centering
  \includegraphics[width=1\linewidth]{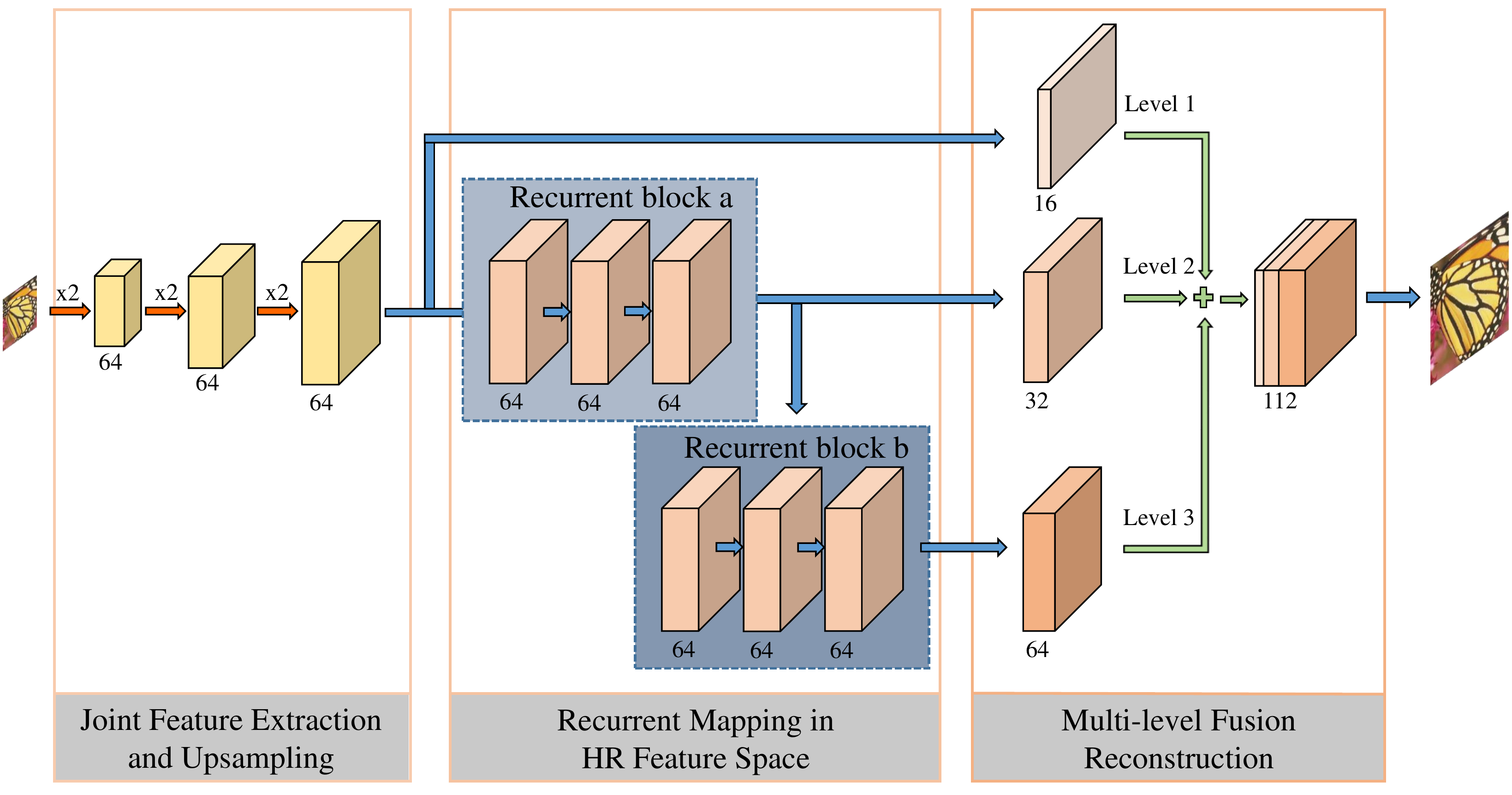}
  \caption{\label{fig:frame}
           \footnotesize{DRFN architecture. Orange arrows indicate transposed convolutions; blue arrows indicate convolutional layers; and the green plus sign indicates the feature maps concatenation operation. Black numbers indicate the number of feature maps; $\times$2 indicates enlargement of the image size by two times.}}
\end{figure*}

However, as the upscaling factor becomes higher, these methods exhibit strong visual artifacts (Figure \ref{fig:xiaoguo}) caused by their network design philosophy. Current approaches possess three inherent limitations. First, most existing methods \cite{dong2016image}\cite{kim2016accurate}\cite{kim2016deeply} apply interpolation strategies such as bicubic interpolation to first process the input image to the desired size and then use CNNs to extract features and learn LR/HR mapping relations. This pre-processing step often results in visible reconstruction artifacts. Second, several methods extract raw features directly from input LR images and replace the pre-defined upsampling operator with transposed convolution \cite{dong2016accelerating} or sub-pixel convolution \cite{shi2016real}. These methods, however, use relatively small networks and cannot learn complicated mapping well due to a limited network capacity. Moreover, these approaches reconstruct HR images in one upsampling step at the end of the network, which increases the difficulties of training for large scaling factors (e.g., $\times$8). Third, in the reconstruction stage, many algorithms have only one reconstruction level and cannot fully leverage more underlying information, including original and complementary information, among different recovery stages. Additionally, images reconstructed by a single-level structure lack many realistic texture details.

To address the above limitations, in this paper, we propose a deep recurrent fusion framework (DRFN) for the large-factor SR problem. As illustrated in Figure \ref{fig:frame}, we jointly extract and upsample raw features from an input LR image by putting the transposed convolution in front of the network. This design does not require a pre-defined upsampling operator (e.g., bicubic interpolation) as the pre-processing step and allows the following convolutional layers to focus on mapping in the HR feature space. After that, we use recurrent residual blocks to gradually recover high-frequency information of the HR image using fewer parameters. Then, three convolutional layers are used to extract features with different receptive field sizes at each recovery stage. In doing so, we can make full use of complementary information among three different level features. Finally, we use a convolutional layer to fuse feature maps and reconstruct HR images.

In summary, we propose a novel DRFN end-to-end framework for single-image super-resolution with high upscaling factors ($\times$4 and $\times$8). Without extraneous steps, this DRFN training from scratch can produce HR images with more texture details and better visual performance. As demonstrated through extensive experiments, the proposed DRFN significantly outperforms existing deep learning methods in terms of accuracy and visual effects, especially when dealing with large scaling factors.

\section{Related Work}
Extensive research has investigated the SR problem. In this section, we summarize the main related works with respect to conventional methods, learning-based methods, and deep-learning-based methods.

\textbf{Conventional Methods.}
Early methods were mainly based on image interpolation, namely linear, bicubic, or Lanczos \cite{duchon1979lanczos}. Later, prior information was introduced to promote results, such as edge prior \cite{dai2007soft} and edge statistics \cite{fattal2007image}. Michaeli et al. \cite{michaeli2013nonparametric} utilized the recurrent property of image patches to recover an SR blur kernel. Efrat et al. \cite{efrat2013accurate} combined accurate reconstruction constraints and gradient regularization to improve reconstruction results. Although most conventional approaches are fast and generate smooth HR images, high-frequency information is difficult to recover as overly smooth solutions.

\textbf{Learning-based Methods.}
More approaches focus on recovering complex mapping relations between LR and HR images. These mapping relations can be established by external or internal databases.

Several methods can learn LR/HR mapping relations from external databases using different models and strategies. Yang et al. \cite{yang2010image} introduced sparse representations of LR/HR patch pairs. Freeman et al. \cite{freeman2002example} presented dictionaries of LR/HR patch pairs and reconstructed HR patches with the corresponding nearest neighbors from the LR space. Timofte et al. \cite{timofte2014a+} assumed all LR/HR patches lie on the manifold in the LR/HR space, so outputs were reconstructed by the retrieved patches. Additionally, K-means \cite{yang2013fast} and random forest \cite{schulter2015fast} algorithms were proposed to seek mapping by partitioning the image database. Methods based on external databases can obtain a mass of different prior knowledge to achieve good performance. Nevertheless, the efficiency of these approaches is rather poor given the cost of matching HR patches.

Methods based on internal databases create LR/HR patch pairs and utilize the self-similarity property of input images. Freedman et al. \cite{freedman2011image} used image pyramids to seek the local self-similarity property. Singh et al. \cite{singh2014super} used directional frequency sub-bands to compose patches. Cui et al. \cite{cui2014deep} conducted image SR layer by layer. In each layer, they elaborately integrated the non-local self-similarity search and collaborative local auto-encoder. Huang et al. \cite{huang2015single} warped the LR patch to find matching patches in the LR image and unwarped the matching patch as the HR patch. Methods based on internal databases have high computational costs to search patches, resulting in slow speed.

Liu et al. \cite{liu2017retrieval} proposed a group-structured sparse representation approach to make full use of internal and external dependencies to facilitate image SR. Xu et al. \cite{xu2018efficient} proposed an integration model based on Gaussian conditional random fields, which learns the probabilistic distribution of the interaction between patch-based and deep-learning-based SR methods.

\textbf{Deep-learning-based Methods.}
Deep learning methods have achieved great success with SR. Dong et al. \cite{dong2016image} successfully pioneered a CNN to solve the SR problem. Shi et al. \cite{shi2016real} presented an efficient sub-pixel convolution (ESPCN) layer to upscale LR feature maps into HR output. By doing so, ESPCN achieves a stunning average speed. Dong et al. \cite{dong2016accelerating} presented a hourglass-shaped CNN to accelerate SRCNN. Motivated by SRCNN, Kim et al. \cite{kim2016accurate} presented a very deep convolutional network (VDSR) to obtain a larger receptive field. VDSR achieves fast convergence via proposed residual learning and gradient clipping. Moreover, VDSR can handle multi-scale SR using a single network. Because deeper networks often introduce more parameters, recurrent learning strategies were applied in this study to reduce the number of parameters along with skip connections to accelerate convergence.

Kim et al. \cite{kim2016deeply} presented an approach that used more layers to increase the receptive field of the network and proposed a very deep recursive layer to avoid excessive parameters. Zhang et al. \cite{zhang2016ccr} proposed an effective and fast SISR algorithm by combining clustering and collaborative representation. Tai et al. introduced recursive blocks in DRRN \cite{Tai-DRRN-2017} and memory blocks in Memnet \cite{Tai-MemNet-2017} for deeper networks, but each method must interpolate the original LR image to the desired size. Yang et al. \cite{yang2017deep} utilized the LR image and its edge map to infer sharp edge details of an HR image during the recurrent recovery process. Lai et al. \cite{LapSRN} presented a Laplacian pyramid network for SR. The proposed model can predict high-frequency information with coarse feature maps. LapSRN is accurate and fast for SISR. Unlike most deep-learning methods, we adopted transposed convolution to replace bicubic interpolation to extract raw features and promote reconstruction performance. Furthermore, a multi-level structure  was designed to obtain better reconstruction performance, including visual effects.
%-------------------------------------------------------------------------
\section{Methodology}
In this section, we describe the design methodology of our proposed DRFN. Figure \ref{fig:frame} presents the DRFN for image reconstruction, which consists of three main parts: joint feature extraction and upsampling, recurrent mapping in the HR feature space, and multi-level fusion reconstruction. First, the proposed method uses multiple transposed convolution operations to jointly extract and upsample raw features from the input image. Second, two recurrent residual blocks are used for mapping in the HR feature space. Finally, DRFN uses three convolutional layers to extract features with different receptive field sizes at each recovery stage and uses one convolutional layer for multi-level fusion reconstruction. Now, we will present additional technical details of each part of our model.

\begin{figure*}[htbp]
  \centering
  \includegraphics[width=.8\linewidth]{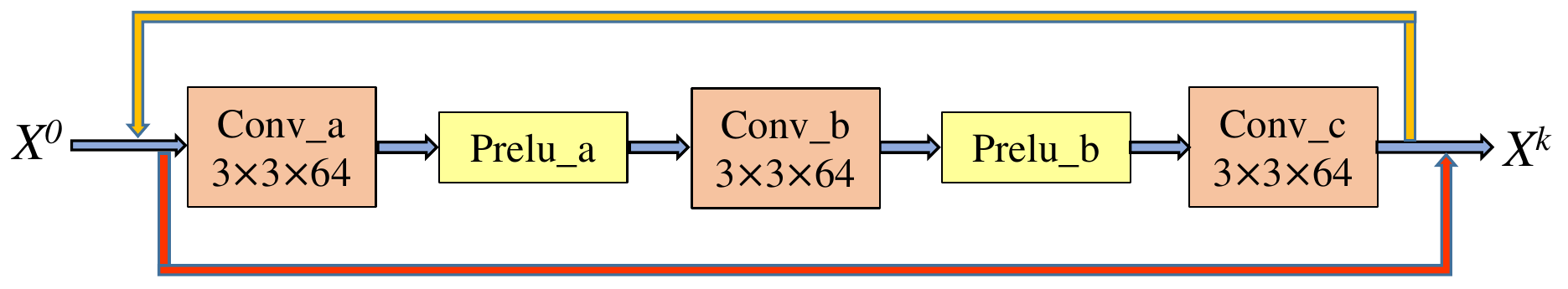}
  \caption{\label{fig:RU}
           \footnotesize{Structure of recurrent residual block. The red line indicates a skip connection, and the orange line indicates a recurrent connection.
           3$\times$3$\times$64 indicates that the size of the convolution kernel is 3$\times$3, and the number of output channels is 64.}}
\end{figure*}

\subsection{Joint Feature Extraction and Upsampling}
In this subsection, we show how to jointly extract and upsample raw features. The transposed convolutional layer consists of diverse, automatically learned upsampling kernels, from which raw features can be extracted simultaneously from the input image to achieve upsampling. For higher SR factors, compared with bicubic interpolation, transposed convolution can alleviate training difficulties while effectively suppressing artifacts. Therefore, we use transposed convolutions to amplify the original LR input image ($X_L$) to the desired size.

FSRCNN \cite{dong2016accelerating} extracts feature maps in the LR space and replaces the bicubic upsampling operation with one transposed convolution layer at the end of the network, and LapSRN progressively reconstructs an HR image throughout the structure. By contrast, the proposed method first puts the transposed convolution at the forefront of the network for joint feature extraction and upsampling. This setting is conducive to allowing the rest of the network to extract features in the HR feature space to further improve performance. Moreover, extracting raw features from the original image enhances the reconstruction details, which is beneficial in generating large-scale and visually satisfying images. The upsampling process can be formulated as
\begin{equation}
\label{formula:upscale}
X_{s+1}=F_p(T_{x2}(X_{s}))~,~~s=0,1,2,\cdots
\end{equation}
where $T_{x2}$ represents the transposed convolution operation that doubles the size of the image, and $F_p$ is a non-linear operation achieved using a parametric rectified linear unit (PReLU). $X_s$ denotes the feature maps extracted from the transposed convolution of step $s$. When $s=0$, $X_s$ is the input LR image $X_L$. Then, the size of the input image is magnified by iterating Eq. \ref{formula:upscale}. The number of iterations can be adjusted to determine the SR scale factor; for example, the scale factor is $8$ when $s=3$.

\subsection{Recurrent Mapping in HR Feature Space}
In this subsection, we show the structure of our recurrent residual block and how recurrent blocks gradually recover high-frequency information in HR images. In ResNet \cite{he2016deep}, the basic residual unit uses batch normalization (BN) \cite{ioffe2015batch} and ReLU as the activation function after the weight layers. BN layers normalize features to limit networks' range of flexibility and occupy considerable GPU memory; therefore, we removed the batch normalization layers from the proposed network as Nah et al. \cite{nah2016deep} did in their image-deblurring work. In addition, we replaced ReLU with parametric ReLU to avoid "dead features" caused by zero gradients in ReLU. Our basic unit for mapping in an HR feature space is illustrated in Figure \ref{fig:RU}. Deep models are prone to overfitting and become disk hungry, hence our adoption of recurrence learning strategies to reduce the number of parameters. Skip connections were used in recurrent blocks to provide fast and improved convergence.

The size of the convolution kernel used for feature extraction was 3, and padding was set to 1 to prevent the size of the feature maps from changing. We used two recurrent residual blocks, each of which looped 10 times. Therefore, our network had a much larger receptive field, which benefits large factors. Different from other methods that map from an LR feature space to HR feature space, the proposed DRFN used recurrent blocks to map in the HR feature space. Output feature maps of the recurrent residual block are progressively updated as follows:
\begin{equation}
X^k=F_c(F_{pb}(F_b(F_{pa}(F_a(X^{k-1})))))+X^{k-1},~k=1,2,3,\cdots
\end{equation}
where $X^k$ represents the $k$-th cyclically generated feature maps; and $F_m(m=a,b,c)$ and $F_{pn}(n=a,b)$ denote convolution and PReLU operation, respectively.

\subsection{Multi-level Fusion Reconstruction}
In this subsection, we show how to fuse different-level features and perform HR image reconstruction. A larger SR factor requires more diverse feature information; to meet this need, we propose fusing different-level features to recover the final HR image. As shown in Figure \ref{fig:frame}, three convolutional layers were used to automatically extract features at different levels. Then, we concatenated these features and ultimately apply one convolutional layer to integrate features with different receptive field sizes. Each recurrent residual block gradually refined the rough image information from the front block but may lose original feature information for reconstruction in this process. Therefore, different-level feature information must be integrated, including refined information and easy-to-lose information (i.e., original feature information), to make full use of complementary information among three different level features. Corresponding experiments demonstrated that fusion networks with three levels improve reconstruction accuracy and visual performance compared with single and double levels, especially for larger-scale factors.

We chose mean square error (MSE) as the loss function. Let $x$ denote the input LR image patch and $y$ indicate the corresponding HR image patch. Given a training dataset $\{x^i, y^i\}_{i=1}^N$ containing $N$ patches, the goal is to minimize the the following formula:
\begin{equation}
L=\frac{1}{2N}\sum_{i=1}^{N}\parallel y^{i}-F_\Theta(x^{i})\parallel ^2
\end{equation}
where $F_\Theta$ represents the feed-forward neural network parameterized by $\Theta$. We used mini-batch stochastic gradient descent (SGD) with backpropagation to optimize the penalty function, and the DRFN was implemented by Caffe \cite{jia2014caffe}.

%-------------------------------------------------------------------------
\section{Experiments}
In this section, we first describe the training and test datasets of our method.
We then introduce the implementation details of the algorithm. Next, we
compare our proposed method with several state-of-the-art SISR methods and
demonstrate the superiority of DRFN. Finally, the contributions of different components are analyzed.

%-------------------------------------------------------------------------
\subsection{Datasets}

\textbf{Training dataset:} RFL \cite{schulter2015fast} and VDSR \cite{kim2016accurate} use a training dataset of 291 images, containing 91
images from Yang et al. \cite{yang2010image} and 200 images from the Berkeley Segmentation Dataset \cite{martin2001database}. We also used 291 images to ensure fair comparison with other methods. In addition, we rotated the original images by $90^{\circ}$, $180^{\circ}$, and $270^{\circ}$ and flipped them horizontally. After this process, each image had eight versions for a total training set of $2,328$ images.

\begin{table*}[htbp]
\begin{center}
\caption{Performance comparison of the proposed method with seven SR algorithms on five benchmarks with scale factors of $\times$2, $\times$3, and $\times$4. Red numbers denote the best performance, and blue numbers denote the second-best performance.}
\begin{tabular}{| l | c | l | l | l | l | l |}

\hline
\multirow{2}{*}{Algorithm} & \multirow{2}{*}{Scale} & \multicolumn{1}{c|}{Set5} &\multicolumn{1}{c|}{Set14} &\multicolumn{1}{c|}{BSDS100} &\multicolumn{1}{c|}{Urban100} &\multicolumn{1}{c|}{ImageNet400} \\
\cline{3-7}
          &  & PSNR~~SSIM~~IFC  & PSNR~~SSIM~~IFC & PSNR~~SSIM~~IFC & PSNR~~SSIM~~IFC & PSNR~~SSIM~~IFC  \\

\hline
\hline
BIC\cite{de1962bicubic}                  &2 & 33.64~0.9293~5.714     & 30.31~0.8693~5.699     & 29.55~0.8432~5.256      & 26.88~0.8409~6.191  &30.03~0.8667~5.970 \\
A+\cite{timofte2014a+}                   &2 & 36.55~0.9545~8.465     & 32.40~0.9064~8.001     & 31.23~0.8868~7.282      & 29.24~0.8944~8.246
&32.05~0.8998~6.913   \\
JOR\cite{dai2015jointly}                 &2 & 36.58~0.9543~8.511     & 32.38~0.9063~8.052     & 31.22~0.8867~7.321      & 29.25~0.8951~8.301
&32.05~0.8998~6.969   \\
SRCNN\cite{dong2016image}                &2 & 36.35~0.9521~7.522     & 32.29~0.9046~7.227     & 31.15~0.8851~6.653      & 29.10~0.8900~7.446
&31.98~0.8970~6.374   \\
FSRCNN\cite{dong2016accelerating}        &2 & 37.00~0.9557~8.047     & 32.75~0.9095~7.727     & 31.51~0.8910~7.068      & 29.88~0.9015~8.005
&32.52~0.9031~6.712   \\
VDSR\cite{kim2016accurate}               &2 &
\color{blue}37.53~\color{blue}0.9587~\color{blue}8.580     &
\color{blue}33.15~\color{blue}0.9132~\color{blue}8.159     &
\color{blue}31.90~\color{blue}0.8960~\color{blue}7.494     &
\color{blue}30.77~\color{blue}0.9143~\color{blue}8.605     &
\color{blue}33.22~\color{blue}0.9106~\color{blue}7.096    \\
LapSRN\_x2\cite{LapSRN}                  &2 & 37.44~0.9581~8.400    & 33.06~0.9122~8.011      & 31.78~0.8944~7.293 &   30.39~0.9096~8.430
&32.98~0.9082~6.912 \\
\textbf{DRFN\_x2}                        &2 & \color{red}37.71~\color{red}0.9595~\color{red}8.927      & \color{red}33.29~\color{red}0.9142~\color{red}8.492      & \color{red}32.02~\color{red}0.8979~\color{red}7.721 & \color{red}31.08~\color{red}0.9179~\color{red}9.076 &
\color{red}33.42~\color{red}0.9123~\color{red}8.002    \\

\hline
\hline
BIC\cite{de1962bicubic}                  &3 & 30.39~0.8673~3.453     & 27.62~0.7756~3.327     & 27.20~0.7394~3.003      & 24.46~0.7359~3.604
&27.91~0.7995~3.363  \\
A+\cite{timofte2014a+}                   &3 & 32.59~0.9077~4.922     & 29.24~0.8208~4.491     & 28.30~0.7844~3.971      & 26.05~0.7984~4.812
&29.42~0.8351~4.034   \\
JOR\cite{dai2015jointly}                 &3 & 32.55~0.9067~4.892     & 29.19~0.8204~4.485     & 28.27~0.7837~3.966      & 25.97~0.7972~4.766
&29.34~0.8343~4.028   \\
SRCNN\cite{dong2016image}                &3 & 32.39~0.9026~4.315     & 29.11~0.8167~4.027     & 28.22~0.7809~3.608      & 25.87~0.7889~4.240
&29.27~0.8294~3.617   \\
FSRCNN\cite{dong2016accelerating}        &3 & 33.16~0.9132~4.963     & 29.55~0.8263~4.551     & 28.52~0.7901~4.025      & 26.43~0.8076~4.841
&29.78~0.8376~4.078   \\
VDSR\cite{kim2016accurate}               &3 & 33.66~\color{blue}0.9213~\color{blue}5.203     &
\color{blue}29.88~\color{blue}0.8330~\color{blue}4.692     &
\color{blue}28.83~\color{blue}0.7976~\color{blue}4.151     &
\color{blue}27.14~\color{blue}0.8284~\color{blue}5.163     &
\color{blue}30.37~\color{blue}0.8509~\color{blue}4.251     \\
LapSRN\_x4\footnotemark[1]\cite{LapSRN}                     &3 & \color{blue}33.78~\color{black}0.9209~5.079      & 29.87~0.8328~4.552      & 28.81~0.7972~3.946 &    27.06~0.8269~5.019 &30.32~0.8497~4.085 \\
\textbf{DRFN\_x3}                        &3 & \color{red}34.01~\color{red}0.9234~\color{red}5.421      & \color{red}30.06~\color{red}0.8366~\color{red}4.897      & \color{red}28.93~\color{red}0.8010~\color{red}4.281      &  \color{red}27.43~\color{red}0.8359~\color{red}5.481      &
\color{red}30.59~\color{red}0.8539~\color{red}4.582     \\

\hline
\hline
BIC\cite{de1962bicubic}              &4    &28.42~0.8099~2.342     & 26.00~0.7025~2.259     & 25.96~0.6692~2.021      & 23.15~0.6592~2.355      & 26.70~0.7530~2.137    \\
A+\cite{timofte2014a+}               &4  &30.28~0.8587~3.248     & 27.32~0.7497~2.962     & 26.82~0.7100~2.551      & 24.34~0.7201~3.180      & 27.92~0.7877~2.660    \\
JOR\cite{dai2015jointly}             &4    &30.19~0.8563~3.190     & 27.27~0.7479~2.923     & 26.79~0.7083~2.534      & 24.29~0.7181~3.113      & 27.87~0.7865~2.630    \\
SRCNN\cite{dong2016image}            &4    &30.48~0.8618~2.991     & 27.50~0.7517~2.751     & 26.90~0.7115~2.396      & 24.16~0.7066~2.769      & 28.18~0.7903~2.492    \\
FSRCNN\cite{dong2016accelerating}    &4    &30.70~0.8646~2.986     & 27.59~0.7539~2.707     & 26.96~0.7174~2.359      & 24.62~0.7281~2.907      & 28.16~0.7895~2.412    \\
VDSR\cite{kim2016accurate}           &4    &31.35~0.8838~\color{blue}3.542      & 28.02~0.7678~\color{blue}3.106      & 27.29~0.7252~\color{blue}2.679 & 25.18~0.7534~\color{blue}3.462      & 28.77~0.8056~\color{blue}2.820  \\
LapSRN\_x4\cite{LapSRN}     &4                 &\color{blue}31.54~\color{blue}0.8852~\color{black}3.515      & \color{blue}28.19~\color{blue}0.7716~\color{black}3.089      & \color{blue}27.32~\color{blue}0.7275~\color{black}2.618 &                                          \color{blue}25.21~\color{blue}0.7554~\color{black}3.448      & \color{blue}28.82~\color{blue}0.8082~\color{black}2.785  \\
\textbf{DRFN\_x4}      &4                   &\color{red}31.55~\color{red}0.8861~\color{red}3.693      & \color{red}28.30~\color{red}0.7737~\color{red}3.250      & \color{red}27.39~\color{red}0.7293~\color{red}2.766 &                                            \color{red}25.45~\color{red}0.7629~\color{red}3.693      & \color{red}28.99~\color{red}0.8106~\color{red}2.954  \\

\hline
\end{tabular}
\label{tab:x234}
\end{center}
\end{table*}

\begin{table*}[htbp]
\begin{center}
\caption{Performance comparison of the proposed method with six SR algorithms on three benchmarks with a scale factor of $\times$8. Red numbers denote the best performance, and blue numbers denote the second-best performance. }
\begin{tabular}{| l | c c c | c c c | c c c |}

\hline
\multirow{2}{*}{Upscaling Factor $\times$8} &\multicolumn{3}{c|}{Set5} &\multicolumn{3}{c|}{Set14} &\multicolumn{3}{c|}{BSDS100}  \\
\cline{2-10}
            & PSNR & SSIM & IFC & PSNR & SSIM & IFC & PSNR & SSIM & IFC   \\

\hline
\hline
BIC\cite{de1962bicubic}                  &24.39&0.657&0.836     &23.19&0.568&0.784     &23.67&0.547&0.646          \\
A+\cite{timofte2014a+}                   &25.52&0.692&1.007     &23.98&0.597&0.983     &24.20&0.568&0.797          \\
SRCNN\cite{dong2016image}                &25.33&0.689&0.938     &23.85&0.593&0.865     &24.13&0.565&0.705          \\
FSRCNN\cite{dong2016accelerating}        &25.41&0.682&0.989     &23.93&0.592&0.928     &24.21&0.567&0.772          \\
VDSR\cite{kim2016accurate}               &25.72&0.711&1.123     &24.21&0.609&1.016     &24.37&0.576&0.816          \\
LapSRN\_x8\cite{LapSRN}                      &\color{blue}{26.14}&\color{blue}0.738&\color{blue}1.295      & \color{blue}24.44&\color{blue}0.623&\color{blue}1.123      & \color{blue}24.54&\color{blue}0.586&\color{blue}0.880       \\
\textbf{DRFN\_x8}                         &\color{red}{26.22}&\color{red}0.740&\color{red}1.331      & \color{red}24.57&\color{red}0.625&\color{red}1.138      & \color{red}24.60&\color{red}0.587&\color{red}0.893      \\

\hline
\end{tabular}
\label{tab:x8}
\end{center}
\end{table*}

\textbf{Test dataset:} We evaluated the results of five test sets---Set5 \cite{bevilacqua2012low}, Set14 \cite{zeyde2010single}, BSDS100 \cite{arbelaez2011contour}, Urban100\cite{huang2015single}, and ImageNet400---which contained $5$, $14$, $100$, $100$, and $400$ images, respectively. In these datasets, Set5, Set14, and BSDS100 were composed of natural scenes and have been used often in other studies. Urban100 was created by Huang et al. \cite{huang2015single} and includes 100 images of various real-world structures, which presents challenges for many methods. Images in ImageNet400 are randomly selected from ImageNet \cite{deng2009imagenet}.

\footnotetext[1]{Due to the network design of LapSRN, scale factors for training are limited to the power of 2 (e.g., $\times$2, $\times$4, or $\times$8). LapSRN performs SR to other scales by first upsampling input images to a larger scale and then downsampling the output to the desired resolution. As mentioned in their paper, we tested the results for $\times$3 SR by using their $\times$4 model.}

%----------------------------------------{}---------------------------------
\subsection{Implementation Details}
We converted original-color RGB images into grayscale images and performed
training and testing on the luminance channel. We generated LR training images using the bicubic downsampling and cut them into patches with a stride of four. Approximately $930,000$ patches were generated after this operation. We set the mini-batch size of SGD with momentum to 32, such that each epoch contained $29,067$ iterations for training. In addition, we set the momentum parameter to 0.9 and weight decay to $10^{-4}$.

All PReLUs were initially set to 0.33. The stride of transposed convolution was $2$ to ensure that each transposed convolution would magnify the image twice. The kernel size of each convolution was $3\times3$. We used the same strategy as He et al. \cite{he2015delving} for convolution weight initialization. The initial learning rate was set to 0.1 and then reduced by a factor of 10 every 10 epochs. We also adopted adjustable gradient clipping \cite{kim2016accurate} to ease the difficulty of training the network. The gradient of each iteration update was limited to $[-\frac{A}{\alpha},\frac{A}{\alpha}]$, where $A=0.01$ is the maximum value of each update step size and $\alpha$ is the current learning rate. We stopped training when the loss ceased to fall. It took approximately 3 days to train the network using an NVIDIA GTX $1080$Ti graphics card.

\begin{figure*}[htbp]
  \begin{minipage}{0.485\linewidth}
  \centerline{\includegraphics[width=1\textwidth]{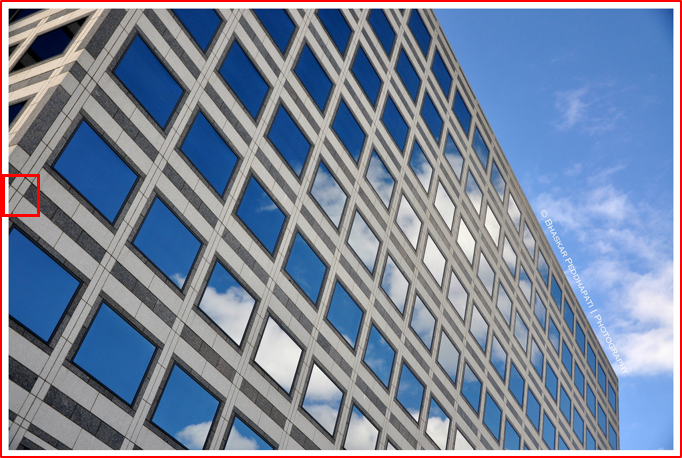}}
  \centerline{Ground Truth}
  \centerline{(GT)}
  \end{minipage}
  \begin{minipage}{0.5\linewidth}
    \begin{minipage}{0.24\linewidth}
    \centerline{\includegraphics[width=1\textwidth]{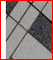}}
    \centerline{GT}
    \centerline{(PSNR,SSIM)}
    \end{minipage}
    \begin{minipage}{0.24\linewidth}
    \centerline{\includegraphics[width=1\textwidth]{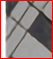}}
    \centerline{Ours}
    \centerline{(\color{red}{28.48}\color{black}{,}\color{red}0.8605\color{black}{)}}
    \end{minipage}
    \begin{minipage}{0.24\linewidth}
    \centerline{\includegraphics[width=1\textwidth]{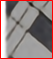}}
    \centerline{LapSRN\cite{LapSRN}}
    \centerline{(27.83,0.8442)}
    \end{minipage}
    \begin{minipage}{0.24\linewidth}
    \centerline{\includegraphics[width=1\textwidth]{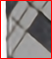}}
    \centerline{VDSR\cite{kim2016accurate}}
    \centerline{(28.03,0.8458)}
    \end{minipage}

    \begin{minipage}{0.24\linewidth}
    \centerline{\includegraphics[width=1\textwidth]{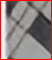}}
    \centerline{FSRCNN\cite{dong2016accelerating}}
    \centerline{(26.78,0.8059)}
    \end{minipage}
    \begin{minipage}{0.24\linewidth}
    \centerline{\includegraphics[width=1\textwidth]{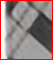}}
    \centerline{SRCNN\cite{dong2016image}}
    \centerline{(26.27,0.7859)}
    \end{minipage}
    \begin{minipage}{0.24\linewidth}
    \centerline{\includegraphics[width=1\textwidth]{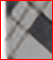}}
    \centerline{A$+$\cite{timofte2014a+}}
    \centerline{(26.65,0.8034)}
    \end{minipage}
    \begin{minipage}{0.24\linewidth}
    \centerline{\includegraphics[width=1\textwidth]{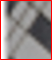}}
    \centerline{Bicubic\cite{de1962bicubic}}
    \centerline{(24.76,0.7220)}
    \end{minipage}
  \end{minipage}
  \caption{ \label{fig:x4-1}
            \footnotesize{Visual comparisons between different algorithms for Urban100 \cite{huang2015single} image with scale factor $\times$4.}}
\end{figure*}

\begin{figure*}[htbp]
  \begin{minipage}{0.44\linewidth}
  \centerline{\includegraphics[width=1\textwidth]{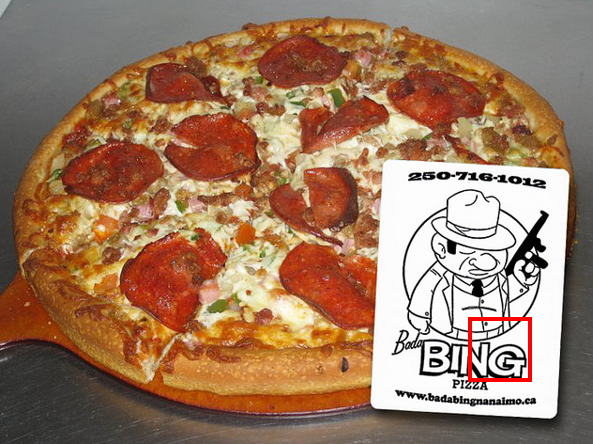}}
  \centerline{Ground Truth}
  \centerline{(GT)}
  \end{minipage}
  \begin{minipage}{0.55\linewidth}
    \begin{minipage}{0.24\linewidth}
    \centerline{\includegraphics[width=1\textwidth]{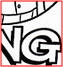}}
    \centerline{GT}
    \centerline{(PSNR,SSIM)}
    \end{minipage}
    \begin{minipage}{0.24\linewidth}
    \centerline{\includegraphics[width=1\textwidth]{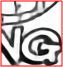}}
    \centerline{Ours}
    \centerline{(\color{red}{24.53}\color{black}{,}\color{red}0.7907\color{black}{)}}
    \end{minipage}
    \begin{minipage}{0.24\linewidth}
    \centerline{\includegraphics[width=1\textwidth]{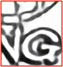}}
    \centerline{LapSRN\cite{LapSRN}}
    \centerline{(24.00,0.7838)}
    \end{minipage}
    \begin{minipage}{0.24\linewidth}
    \centerline{\includegraphics[width=1\textwidth]{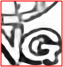}}
    \centerline{VDSR\cite{kim2016accurate}}
    \centerline{(24.21,0.7856)}
    \end{minipage}

    \begin{minipage}{0.24\linewidth}
    \centerline{\includegraphics[width=1\textwidth]{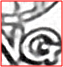}}
    \centerline{FSRCNN\cite{dong2016accelerating}}
    \centerline{(23.33,0.7630)}
    \end{minipage}
    \begin{minipage}{0.24\linewidth}
    \centerline{\includegraphics[width=1\textwidth]{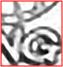}}
    \centerline{SRCNN\cite{dong2016image}}
    \centerline{(22.74,0.7474)}
    \end{minipage}
    \begin{minipage}{0.24\linewidth}
    \centerline{\includegraphics[width=1\textwidth]{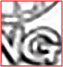}}
    \centerline{A$+$\cite{timofte2014a+}}
    \centerline{(21.67,0.7524)}
    \end{minipage}
    \begin{minipage}{0.24\linewidth}
    \centerline{\includegraphics[width=1\textwidth]{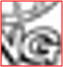}}
    \centerline{Bicubic\cite{de1962bicubic}}
    \centerline{(21.60,0.7016)}
    \end{minipage}
  \end{minipage}
  \caption{ \label{fig:x4-2}
            \footnotesize{Visual comparisons between different algorithms for ImageNet400 image with scale factor $\times$4.}}
\end{figure*}

%-------------------------------------------------------------------------
\subsection{Benchmark Results}
\textbf{Quantitative Evaluation.} Here, we provide quantitative comparisons for $\times$2, $\times$3, and $\times$4 SR results in Table \ref{tab:x234} and $\times$8 SR results in Table \ref{tab:x8}, respectively. We compare our proposed method with bicubic interpolation and the following six state-of-the-art SR methods: A+ \cite{timofte2014a+}, JOR \cite{dai2015jointly}, SRCNN \cite{dong2016image}, FSRCNN \cite{dong2016accelerating}, VDSR \cite{kim2016accurate}, and LapSRN \cite{LapSRN}. We evaluated SR images based on three commonly used image quality metrics: peak signal-to-noise ratio (PSNR), structural similarity (SSIM) \cite{wang2004image}, and information fidelity criterion (IFC) \cite{sheikh2005information}. In particular, IFC has been shown to be related to human visual perception \cite{yang2014single}. For fair comparison of the $\times$8 factor, we followed LapSRN \cite{LapSRN} in using their datasets generated by retrained models of A+ \cite{timofte2014a+}, SRCNN \cite{dong2016image}, FSRCNN \cite{dong2016accelerating}, and VDSR \cite{kim2016accurate}. For scale factors of four and eight, our approach was superior to other SR methods on all datasets.

\begin{figure*}[htbp]
  \begin{minipage}{0.328\linewidth}
  \centerline{\includegraphics[width=1\textwidth]{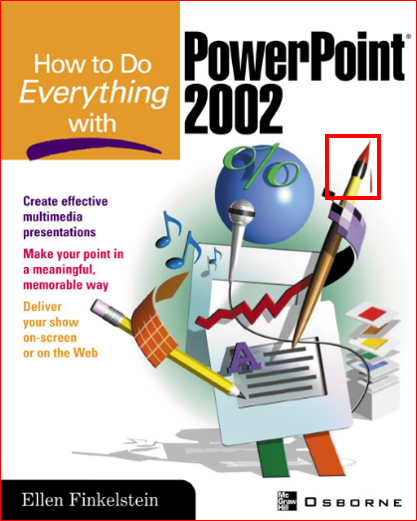}}
  \centerline{Ground Truth}
  \centerline{(GT)}
  \end{minipage}
  \begin{minipage}{0.77\linewidth}
    \begin{minipage}{0.2\linewidth}
    \centerline{\includegraphics[width=1\textwidth]{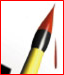}}
    \centerline{GT}
    \centerline{(PSNR,SSIM)}
    \end{minipage}
    \begin{minipage}{0.2\linewidth}
    \centerline{\includegraphics[width=1\textwidth]{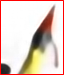}}
    \centerline{Ours}
    \centerline{(\color{red}{20.31}\color{black},\color{red}0.7926\color{black}{)}}
    \end{minipage}
    \begin{minipage}{0.2\linewidth}
    \centerline{\includegraphics[width=1\textwidth]{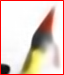}}
    \centerline{LapSRN\cite{LapSRN}}
    \centerline{(20.27,0.7844)}
    \end{minipage}
    \begin{minipage}{0.2\linewidth}
    \centerline{\includegraphics[width=1\textwidth]{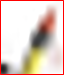}}
    \centerline{VDSR\cite{kim2016accurate}}
    \centerline{(18.73,0.6827)}
    \end{minipage}

    \begin{minipage}{0.2\linewidth}
    \centerline{\includegraphics[width=1\textwidth]{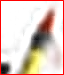}}
    \centerline{FSRCNN\cite{dong2016accelerating}}
    \centerline{(19.45,0.7096)}
    \end{minipage}
    \begin{minipage}{0.2\linewidth}
    \centerline{\includegraphics[width=1\textwidth]{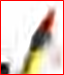}}
    \centerline{SRCNN\cite{dong2016image}}
    \centerline{(19.52,0.7246)}
    \end{minipage}
    \begin{minipage}{0.2\linewidth}
    \centerline{\includegraphics[width=1\textwidth]{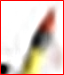}}
    \centerline{A$+$\cite{timofte2014a+}}
    \centerline{(19.48,0.7171)}
    \end{minipage}
    \begin{minipage}{0.2\linewidth}
    \centerline{\includegraphics[width=1\textwidth]{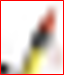}}
    \centerline{Bic\cite{de1962bicubic}}
    \centerline{(18.76,0.6838)}
    \end{minipage}
  \end{minipage}
\caption{ \label{fig:x8-1}
            \footnotesize{Visual comparisons between different algorithms for Set14 \cite{arbelaez2011contour} image with scale factor $\times$8.}}
\end{figure*}

\begin{figure*}[htbp]
  \begin{minipage}{0.224\linewidth}
  \centerline{\includegraphics[width=1\textwidth]{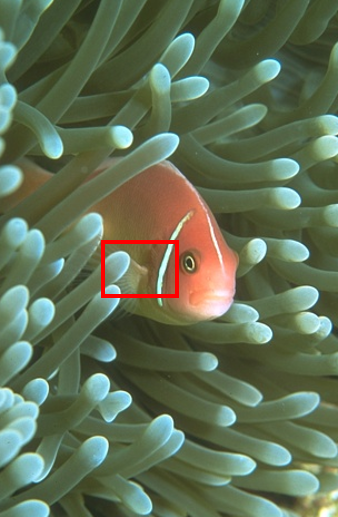}}
  \centerline{Ground Truth}
  \centerline{(GT)}
  \end{minipage}
  \begin{minipage}{0.77\linewidth}
    \begin{minipage}{0.24\linewidth}
    \centerline{\includegraphics[width=1\textwidth]{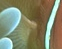}}
    \centerline{GT}
    \centerline{(PSNR,SSIM)}
    \end{minipage}
    \begin{minipage}{0.24\linewidth}
    \centerline{\includegraphics[width=1\textwidth]{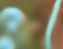}}
    \centerline{Ours}
    \centerline{(\color{red}{27.79}\color{black},\color{red}0.8018\color{black}{)}}
    \end{minipage}
    \begin{minipage}{0.24\linewidth}
    \centerline{\includegraphics[width=1\textwidth]{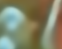}}
    \centerline{LapSRN\cite{LapSRN}}
    \centerline{(27.50,0.7968)}
    \end{minipage}
    \begin{minipage}{0.24\linewidth}
    \centerline{\includegraphics[width=1\textwidth]{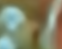}}
    \centerline{VDSR\cite{kim2016accurate}}
    \centerline{(26.68,0.7634)}
    \end{minipage}

    \begin{minipage}{0.24\linewidth}
    \centerline{\includegraphics[width=1\textwidth]{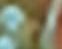}}
    \centerline{FSRCNN\cite{dong2016accelerating}}
    \centerline{(26.44,0.7452)} \end{minipage}
    \begin{minipage}{0.24\linewidth}
    \centerline{\includegraphics[width=1\textwidth]{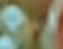}}
    \centerline{SRCNN\cite{dong2016image}} \centerline{(26.40,0.7436)}
    \end{minipage} \begin{minipage}{0.24\linewidth}
    \centerline{\includegraphics[width=1\textwidth]{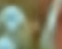}}
    \centerline{A$+$\cite{timofte2014a+}} \centerline{(26.48,0.7544)}
    \end{minipage} \begin{minipage}{0.24\linewidth}
    \centerline{\includegraphics[width=1\textwidth]{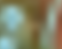}}
    \centerline{Bicubic\cite{de1962bicubic}} \centerline{(25.15,0.7161)}
    \end{minipage}
  \end{minipage}
  \caption{ \label{fig:x8-2}
            \footnotesize{Visual comparisons between different algorithms for BSDS100 \cite{arbelaez2011contour} with scale factor $\times$8.}}
\end{figure*}

\textbf{Visual Performance.}
As indicated in Tables \ref{tab:x234} and \ref{tab:x8}, DRFN was found to be far superior to other methods in IFC on all datasets. Combined with the visual samples in Figures \ref{fig:x4-1}, \ref{fig:x4-2}, \ref{fig:x8-1}, and \ref{fig:x8-2}, findings show that the proposed DRFN estimated better visual details. For instance, images generated by other SR methods exhibited visible artifacts, whereas the proposed method generated a more visually pleasant image with clean details and sharp edges. For example, in Figure \ref{fig:x4-2}, the results of other methods are completely blurred, and only our result demonstrates clear textures. The experimental results show that the proposed method can achieve good visual performance.

% The picture of pen

%-------------------------------------------------------------------------
\subsection{Model Analysis}
In this subsection, we first compare $\times2$ and $\times3$ SR results of the proposed DRFN with existing methods. Then, we study the contributions of different components of the proposed DRFN to SR reconstruction and explore the effects of cycle times of recurrent blocks on reconstruction performance.

\textbf{Comparisons of $\times2$ and $\times3$ SR Results.} Although our work is intended for for large-factor SR problems, the proposed DRFN can also perform $\times2$ and $\times3$ SR well. The quantitative results for $\times2$ and $\times3$ SR are presented in Table \ref{tab:x234}. The proposed DRFN significantly outperformed existing methods on $\times2$ and $\times3$ SR results, suggesting that our methodology is reasonable and effective. This DRFN is hence powerful enough to handle different scaling factors.

\begin{table}[htbp]
\begin{center}
\caption{Average PSNR when DRFN performs image magnification using bicubic and transposed convolution at the front and last of the network, respectively, for scale factor $\times4$ and $\times8$ on dataset Set5 \cite{bevilacqua2012low}, Set14 \cite{zeyde2010single}, and BSDS100 \cite{arbelaez2011contour}.}
\begin{tabular}{| l | c | c | c |}

\hline
% \multirow{2}{*}{Versions} &{Set5} &{Set14} &{BSDS100}  \\
% \cline{2-4}
%             & PSNR & PSNR & PSNR    \\
{Versions} &{Set5} &{Set14} &{BSDS100}  \\

\hline
\hline
$\times$4-prebic    & 31.41    &28.10    &27.29\\
$\times$4-posttransconv  & 31.50    &28.19   &27.34\\
$\times$4-pretransconv   & 31.55    &28.30   &27.39\\
\hline
\hline
$\times$8-posttransconv  & 26.03    &24.39    &24.50\\
$\times$8-pretransconv   & 26.22    &24.47     &24.60\\

\hline
\end{tabular}
\label{tab:deconv}
\end{center}
\end{table}

\begin{figure*}[htbp]
  \centering
  \includegraphics[width=1\linewidth]{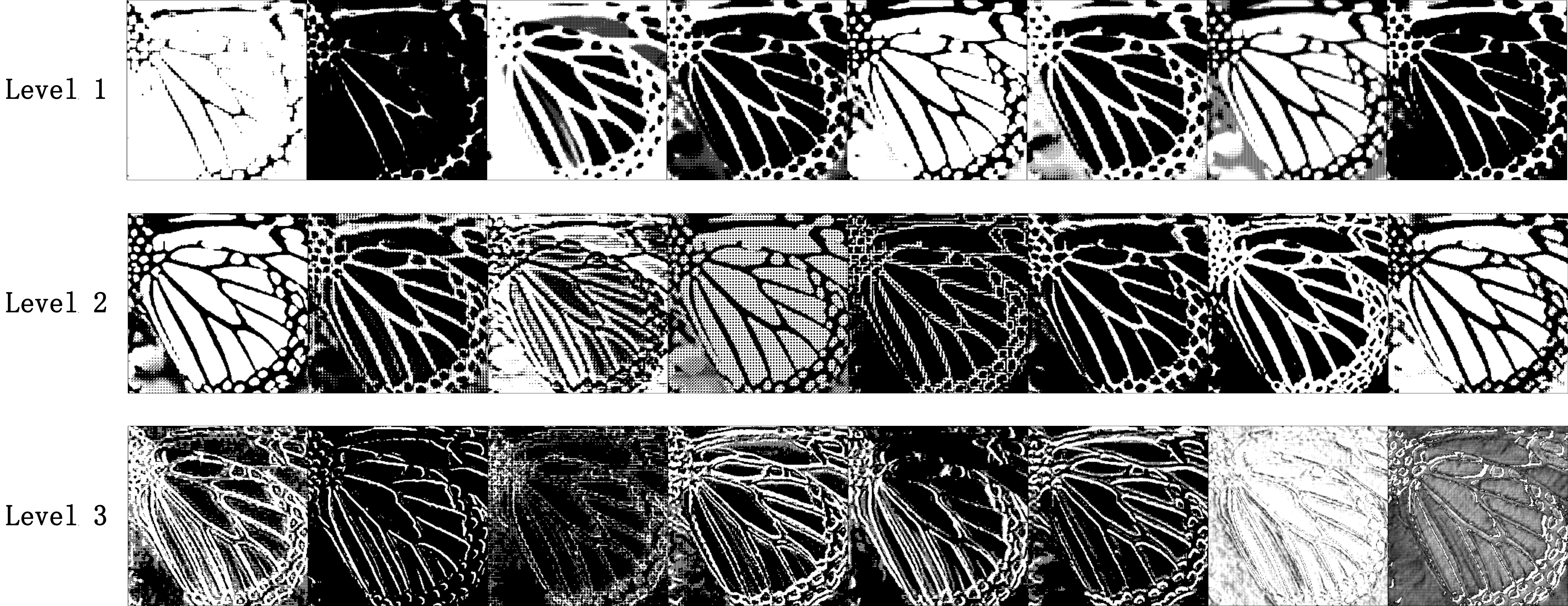}

  \caption{ \label{fig:level}
  \footnotesize{Feature maps of Levels 1, 2, and 3 from Figure \ref{fig:frame} with the image of a ``butterfly" in Set5 as input. The figure shows parts of all feature maps.}}
\end{figure*}

\begin{figure*}[htbp]
  \centering
  \includegraphics[width=1\linewidth]{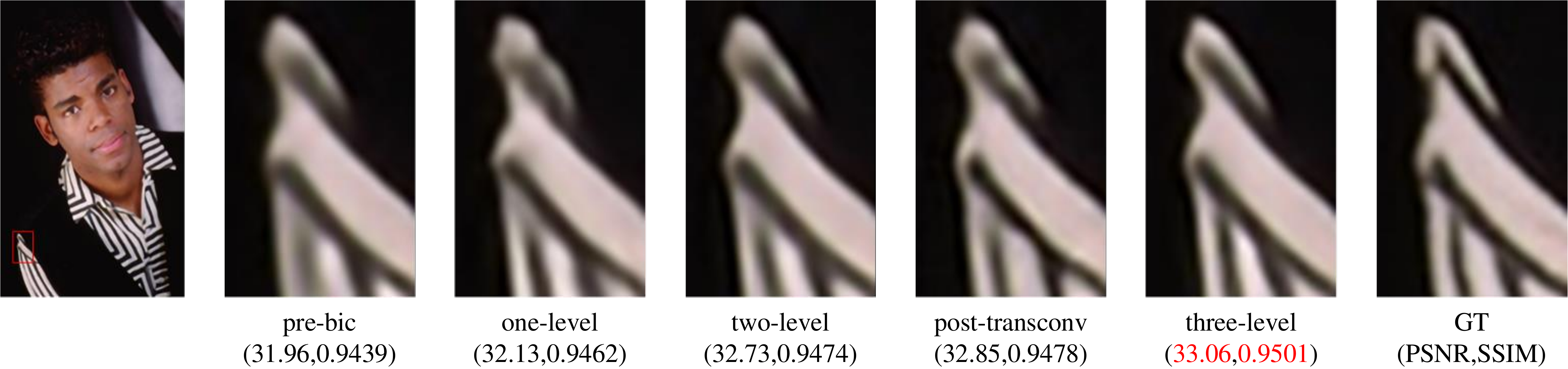}

  \caption{ \label{fig:contrast}
  			\footnotesize{Contribution of different components in the proposed network.}}
\end{figure*}

\begin{figure}[tbp]
  \centering
  \includegraphics[width = 1\linewidth]{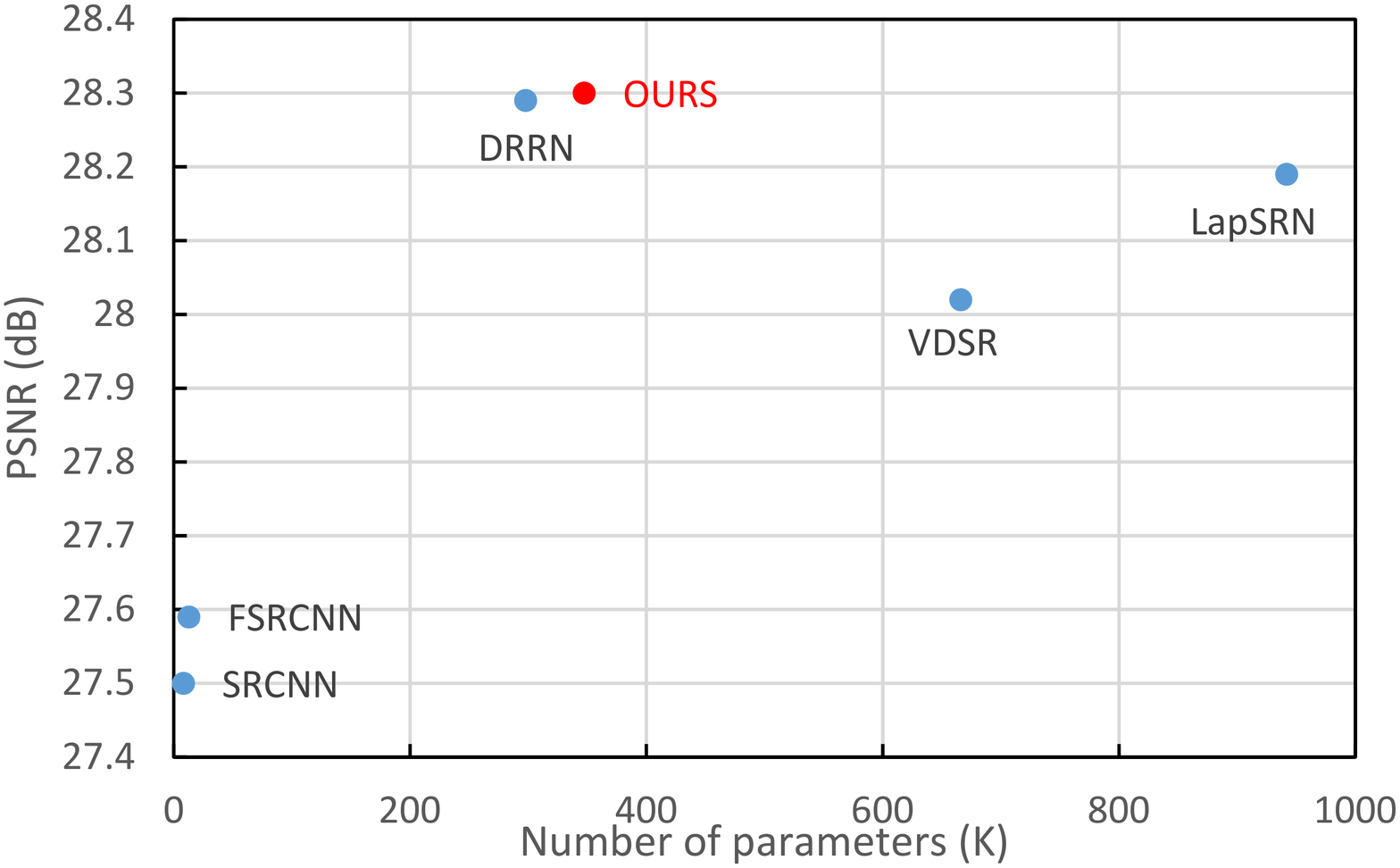}
  \caption{
            \footnotesize{PSNR and parameters of existing CNN models for scale factor $\times$4 on Set14 \cite{zeyde2010single}. Red point is our model. With an appropriate number of parameters, DRFN achieves better performance than state-of-the-art methods.}}
  \label{fig:parameter}
\end{figure}

\textbf{Transposed Convolution.} First, to verify the superiority of transposed convolution compared to bicubic interpolation, we used an interpolated image as input and replaced the transposed convolution with general convolution. Second, we used the original small image as input and placed transposed convolution at the final part of the network to enlarge the image. By doing so, we can prove that the location of transposed convolution has an effect on reconstruction results. We carried out experiments on $\times$4 and $\times$8 scale factors; results are shown in Table \ref{tab:deconv}. In the table, $\times$4-prebic and $\times$4(8)-posttransconv represent the above two mentioned contrast experiments, respectively; and $\times$4(8)-pretransconv is our DRFN\_$\times$4(8). Visual results are shown in Figure \ref{fig:contrast}(the three-level is DRFN\_$\times$4). Quantitative and qualitative results indicate that using transposed convolution to replace bicubic interpolation and placing the transposed convolution at the forefront of the network can boost performance. For example, $\times$4-pretransconv was $0.14dB$ higher than $\times$4-prebic on Set5, and $\times$8-pretransconv was $0.10dB$ higher than $\times$8-posttransconv on BSDS100. Although pretransconv and posttransconv had the same number of model parameters, pretransconv was more computationally intensive in the prediction phase because it caused the following convolution layers to recover high-frequency information in the HR space. The average inference times of post-transconv and pre-transconv for each image in Set5 \cite{bevilacqua2012low} with a scale factor of $\times$4 were 0.584s and 1.187s, respectively. For greater performance improvements, we chose pretransconv as our model.

\begin{table}[tbp]
\begin{center}
\caption{Average PSNR and SSIM of DRFN at different levels, for scale factor $\times4$ and $\times8$ on dataset Set5 \cite{bevilacqua2012low}, Set14 \cite{zeyde2010single}, and BSDS100 \cite{arbelaez2011contour}.}
\begin{tabular}{| c | c | c c | c c | c c |}

\hline
\multirow{2}{*}{Scale} &\multirow{2}{*}{Versions} &\multicolumn{2}{c|}{Set5} &\multicolumn{2}{c|}{Set14} &\multicolumn{2}{c|}{BSDS100} \\
\cline{3-8}
           && PSNR & SSIM & PSNR & SSIM & PSNR & SSIM     \\

\hline
\hline
\multirow{3}{*}{$\times$4}   & one-level & 31.25 &0.879 &28.09&0.767 &27.23 &0.723 \\
                             & two-level & 31.41 &0.883 &28.14&0.770 &27.28 &0.726 \\
                             & three-level & 31.55 &0.886 &28.30&0.774 &27.39 &0.729 \\
\hline
\multirow{3}{*}{$\times$8}   & one-level & 25.75 &0.716 &24.19&0.610 &24.40 &0.578\\
                             & two-level & 25.85 &0.721 &24.28&0.614 &24.45 &0.580\\
                             & three-level & 26.22 &0.740 &24.57&0.625 &24.60 &0.587\\
% \hline
% \multirow{3}{*}{$12\times$}   & single & 23.19 &0.568 &7&0.568 &7 &7\\
%                              & double & 23.19 &0.568 &7&0.568 &7 &7 \\
%                              & three & 23.19 &0.568 &7&0.568 &7 &7 \\

\hline
\end{tabular}
\label{tab:branch}
\end{center}
\end{table}

\textbf{Recurrent residual learning.}
Kim et al. \cite{kim2016accurate} demonstrated that the performance of the network improved as the depth increased. However, deeper networks need to train more parameters. Our use of a recurrent learning strategy greatly reduced the model complexity. For instance, for a recurrent residual block with three convolution layers, looping five times eliminated $(3\times3\times64\times64+64)\times3\times(5-1)=443,136$ parameters. After calculation, our DRFN contained $347,297$ total parameters. Figure \ref{fig:parameter} shows the PSNR performance of several recent CNN models for SR versus the number of parameters. The proposed method achieved better performance with an appropriate number of model parameters.

\begin{table}[htbp]
\begin{center}
\caption{Comparison of different DRFN depths. Different depths can be achieved by setting cycle times of recurrent blocks. Time is the average inference time per image in BSD100 as measured on an NVIDIA GTX 1080Ti GPU.}
% \begin{tabular}{| c | c | c c c |}

% \hline
% \multirow{2}{*}{Cycle Times} &\multirow{2}{*}{Model Size} &\multicolumn{3}{c|}{BSDS100} \\
% \cline{3-5}
%            && PSNR & SSIM & Time(s)  \\

% \hline
% \hline
% 3   & 3.0 MB & 26.67 &0.7046 &0.59 \\
% 5   & 4.9 MB & 27.32 &0.7270 &0.70 \\
% 10  & 9.4 MB & 27.39 &0.7295 &0.86 \\
% \hline
% \end{tabular}
% \label{tab:depth}
% \end{center}
% \end{table}
\begin{tabular}{| l | c c c |}

\hline
\multirow{2}{*}{Cycle Times} &\multicolumn{3}{c|}{BSDS100} \\
\cline{2-4}
           & PSNR & SSIM & Time(s)  \\

\hline
\hline
  \qquad3& 26.67 &0.7046 &0.62 \\
  \qquad5& 27.32 &0.7270 &0.80 \\
  \qquad10& 27.39 &0.7293 &1.28 \\
\hline
\end{tabular}
\label{tab:depth}
\end{center}
\end{table}

\textbf{Multi-level structure.}
To verify that the multi-level structure demonstrated an improved role in image reconstruction, (1) we removed the first two levels from DRFN (denoted as one-level), and (2) removed Level 2 from DRFN (denoted as two-level) for experimental comparison. The result of the three-level network was best as shown in Table \ref{tab:branch}. As displayed in Figure \ref{fig:contrast}, the three-level network reconstructed the image with richer texture details compared with the one-level and two-level networks. Each level had a positive effect on the result. Taking the image ``butterfly" in Set5 as input, feature maps of different levels appear in Figure \ref{fig:level}. These results suggest that the features at each recovery stage had different context characteristics. The multi-level structure rendered our model more robust and more accurate.

\textbf{Network depth.}
We also studies the effect of the cycle times of the recurrent block. Different cycle times indicated that the network had a different depth. We set the number of cycles to $3$, $5$, and $10$, respectively, and the depth of the two recurrent blocks remained the same. We did not continue to train deeper networks due to GPU memory limitations. The experimental results in Table \ref{tab:depth} show that increasing the number of cycles can boost performance but also increases time consumption. To achieve better results, we chose to cycle 10 times as a benchmark.

%-------------------------------------------------------------------------
\section{Conclusions} In this paper, we propose a DRFN for large-scale accurate SISR. Our DRFN uses transposed convolution to jointly extract and upsample raw features, and the following convolution layers focus on mapping in the HR feature
space. High-frequency information is gradually recovered by recurrent residual blocks. Multi-level fusion makes full use of potential information for HR image reconstruction. The proposed DRFN extends quantitative and qualitative SR performance to a new state-of-the-art level. Extensive benchmark experiments and analyses indicate that DRFN is a superior SISR method, especially for large factors.

As this DRFN has achieved outstanding performance on $\times$4 and $\times$8, we intend to apply it to more challenging up-scaling factors such as $\times$12. We also plan to generalize our method for other applications, such as denoising and deblurring.

%\section*{Acknowledgements}
%We thank the anonymous reviewers for the insightful and constructive comments. The work was partially funded by NSFC grant from National Natural Science Foundation of China (NO.91748104, 61632006, 61425002, U1708263).

% or


%\appendices
%% you can choose not to have a title for an appendix
%% if you want by leaving the argument blank
%\section{}
%% use section* for acknowledgment
%\section*{Acknowledgment}

\ifCLASSOPTIONcaptionsoff
  \newpage
\fi



\bibliographystyle{IEEEtran}
\bibliography{tmm}

%\begin{thebibliography}{1}
%\end{thebibliography}


%\noindent{\textbf{Author Biographies:} \vspace{-0.5in}
\newpage
\vspace{-0.6in}
\begin{IEEEbiography}[{\includegraphics[width=0.95in,height=1.25in,clip]{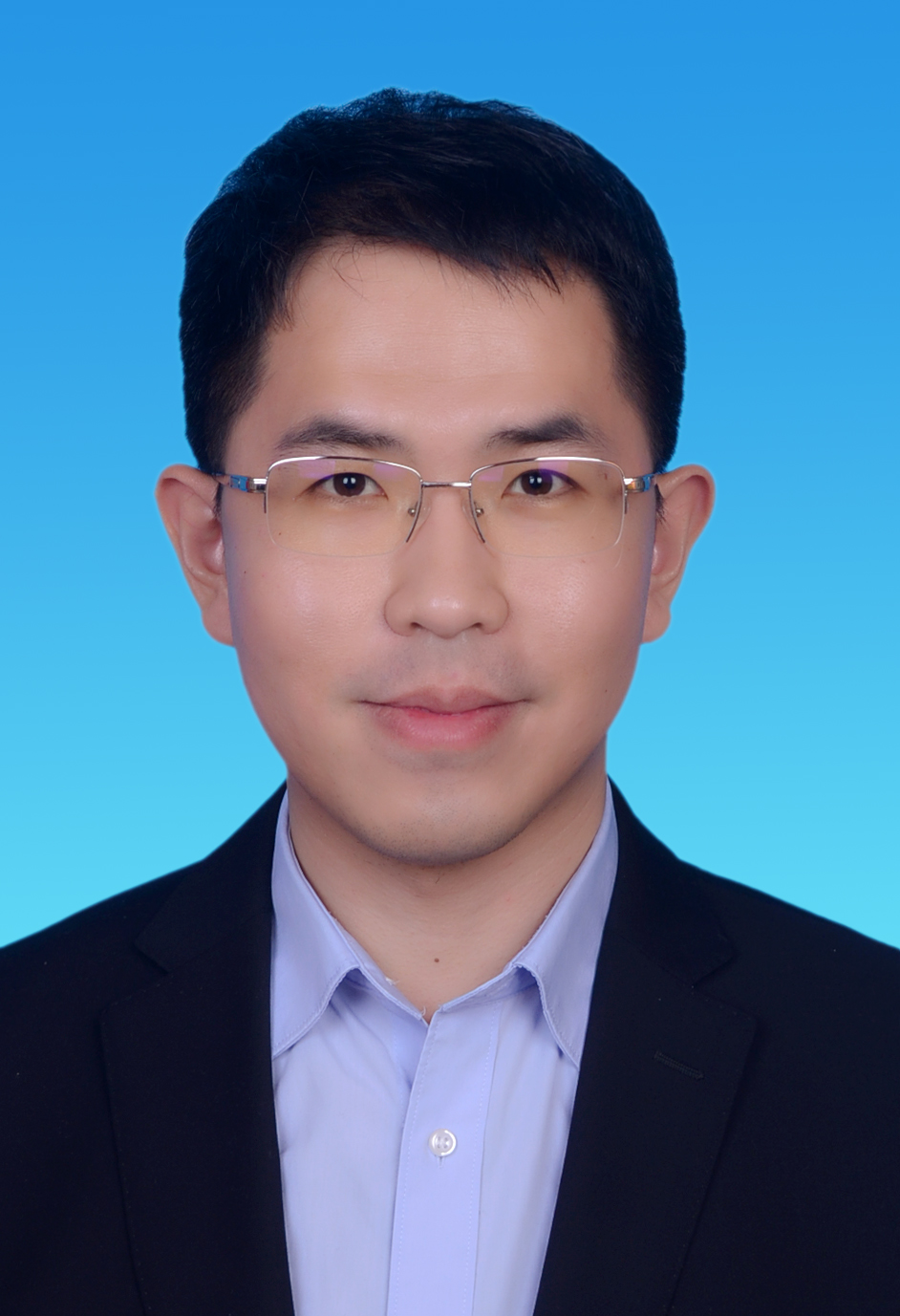}}]{Xin Yang}
is an Associate Professor in the Department of Computer Science at Dalian University of Technology, China. Yang received his B.S. degree in Computer
Science from Jilin University in 2007. From 2007 to June 2012, he was a joint
Ph.D. student at Zhejiang University and UC Davis for Graphics and received
his Ph.D. degree in July 2012. His research interests include computer
graphics and robotic vision.
\end{IEEEbiography}

\vspace{-0.6in}
\begin{IEEEbiography}[{\includegraphics[width=0.95in,height=1.23in,clip]{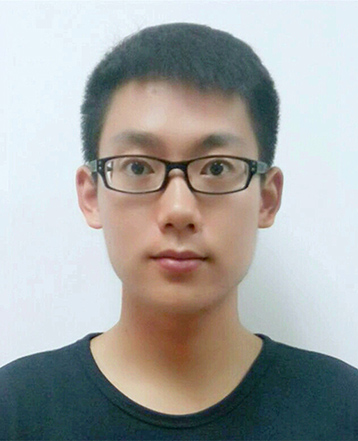}}]{Haiyang Mei}
is a graduate student in the Department of Computer Science at Dalian
University of Technology, China. His research interests include computer
vision, machine learning, and image processing.
\end{IEEEbiography}

\vspace{-0.6in}
\begin{IEEEbiography}[{\includegraphics[width=0.95in,height=1.25in,clip]{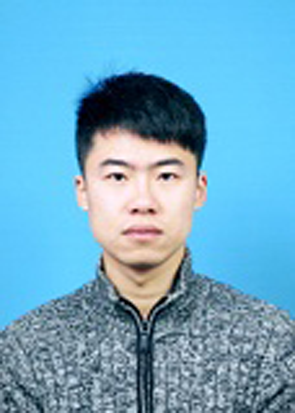}}]{Jiqing Zhang}
is a graduate student in the Department of Computer Science at Dalian
University of Technology, China. His research interests include computer
vision, machine learning, and image processing.
\end{IEEEbiography}

\vspace{-0.6in}
\begin{IEEEbiography}[{\includegraphics[width=0.95in,height=1.25in,clip]{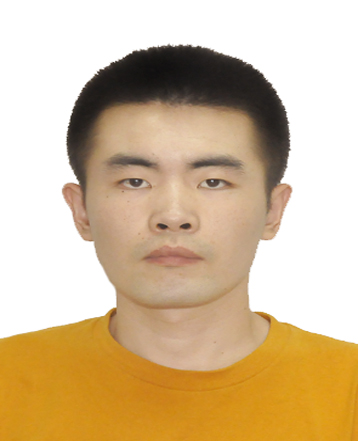}}]{Ke Xu}
is a Ph.D candidate in the Department of Computer Science at Dalian University
of Technology, China. His research interests include computer vision and
machine learning.
\end{IEEEbiography}

\vspace{-0.6in}
\begin{IEEEbiography}[{\includegraphics[width=0.95in,height=1.25in,clip]{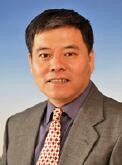}}]{Baocai Yin}
is a Professor of Computer Science at Dalian University of Technology and
the Dean of the Faculty of Electronic Information and Electrical Engineering. His
research concentrates on digital multimedia and computer vision. He received
his B.S. degree and Ph.D. degree in Computer Science, each from Dalian
University of Technology.
\end{IEEEbiography}
\vspace{-0.5in}

\newpage
\begin{IEEEbiography}[{\includegraphics[width=0.95in,height=1.25in,clip]{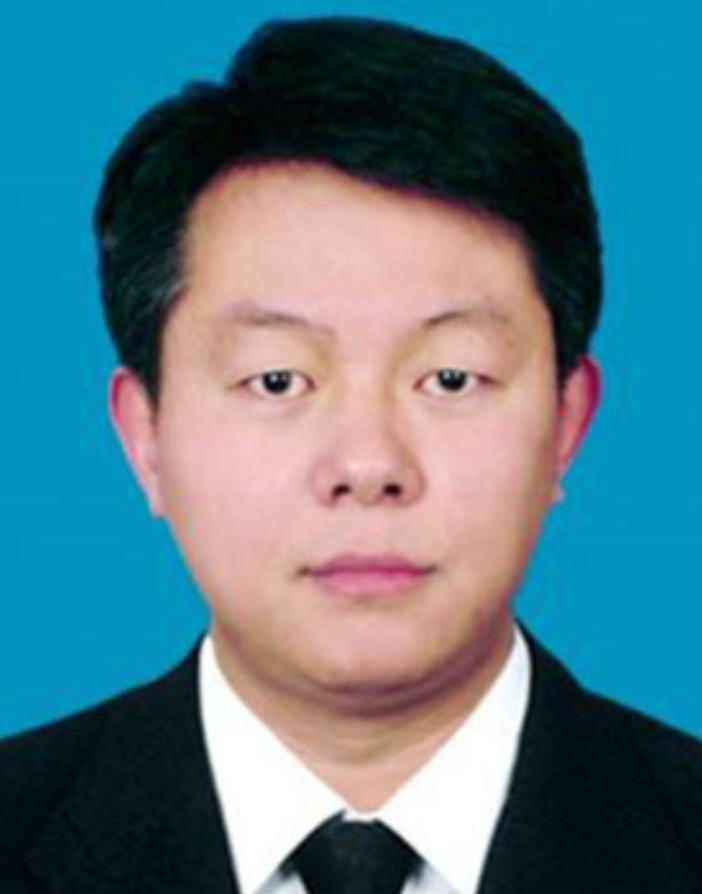}}]{Qiang Zhang}
was born in Xian, China, in 1971. He received his M.Eng. degree in economic
engineering and Ph.D degree in circuits and systems from Xidian
University, Xian, China, in 1999 and 2002, respectively. He was a lecturer
at the Center of Advanced Design Technology, Dalian University, Dalian,
China, in 2003 and was a professor in 2005. His research interests are
bio-inspired computing and its applications. He has authored more than 70
papers in the above fields. Thus far, he has served on the editorial board of
seven international journals and has edited special issues in journals such
as \emph{Neurocomputing} and \emph{International Journal of Computer Applications in
Technology}.
\end{IEEEbiography}
\vspace{-0.6in}

\begin{IEEEbiography}[{\includegraphics[width=0.95in,height=1.25in,clip]{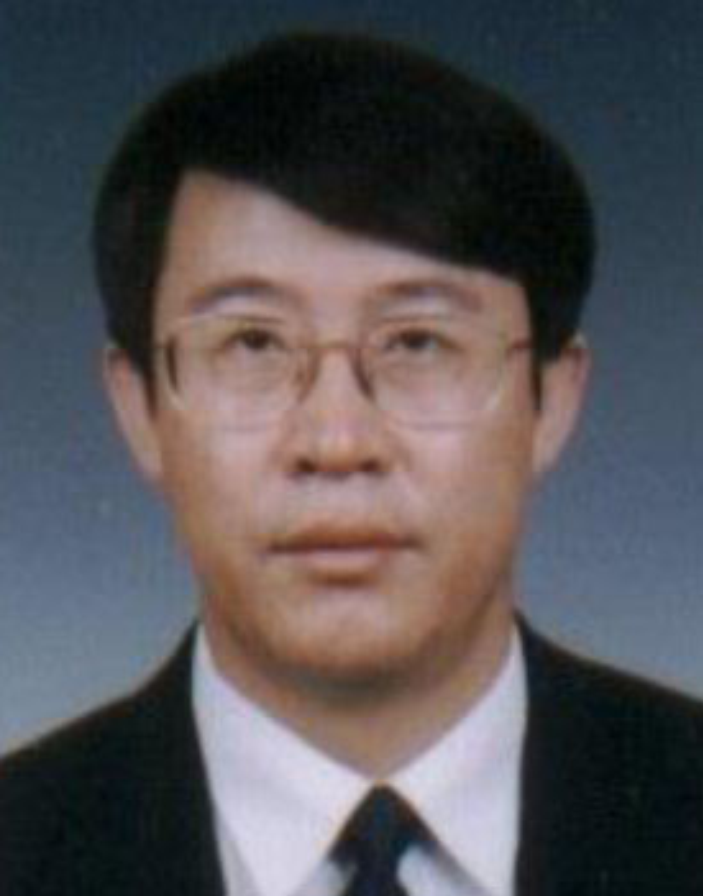}}]{Xiaopeng Wei}
was born in Dalian, China, in 1959. He received his Ph.D degree from Dalian
University of Technology in 1993. He is a professor at Dalian University of
Technology. His research areas include computer animation, computer vision,
robots, and intelligent CAD. So far, he has (co-)authored approximately 200 published papers.
\end{IEEEbiography}
\vspace{4.6in}


%\vfill


\end{document}